\documentclass[10pt,twocolumn,letterpaper]{article}

\usepackage[pagenumbers]{cvpr} %

\usepackage{amsmath}
\usepackage{amssymb}
\usepackage{booktabs}
\usepackage{bbding}
\usepackage{graphicx}
\usepackage{animate}
\usepackage{grffile}
\usepackage{siunitx} %
\usepackage{enumitem}
\usepackage{float}
\usepackage[table, usenames, dvipsnames]{xcolor}
\usepackage{balance}
\usepackage{overpic}

\definecolor{mycyan}{rgb}{0,0.76,0.75}
\definecolor{LightGrey}{rgb}{0.92,0.92,0.92}

\definecolor{cvprblue}{rgb}{0.21,0.49,0.74}
\usepackage[pagebackref,breaklinks,colorlinks,citecolor=cvprblue]{hyperref}

\usepackage[capitalize]{cleveref}
\crefname{section}{Sec.}{Secs.}
\Crefname{section}{Section}{Sections}
\Crefname{table}{Table}{Tables}
\crefname{table}{Tab.}{Tabs.}

\usepackage{times}
\usepackage{epsfig}
\usepackage{amsmath}
\usepackage{amssymb}
\usepackage{booktabs}
\usepackage{makecell}
\usepackage{multirow}
\usepackage{tabularx}
\usepackage{bbm}
\usepackage{caption} 
\usepackage{subcaption}
\usepackage{animate}
\usepackage{pifont}     %
\usepackage{tikz}

\usepackage{xcolor}
\usepackage[bottom]{footmisc}

\def\cvprPaperID{2803} %
\def\confName{CVPR}
\def\confYear{2024}

\definecolor{turquoise}{cmyk}{0.65,0,0.1,0.3}
\definecolor{purple}{rgb}{0.65,0,0.65}
\definecolor{dark_green}{rgb}{0, 0.5, 0}
\definecolor{orange}{rgb}{0.8, 0.6, 0.2}
\definecolor{red}{rgb}{0.8, 0.2, 0.2}
\definecolor{darkred}{rgb}{0.6, 0.1, 0.05}
\definecolor{blueish}{rgb}{0.0, 0.3, .6}
\definecolor{light_gray}{rgb}{0.7, 0.7, .7}
\definecolor{pink}{rgb}{1, 0, 1}
\definecolor{greyblue}{rgb}{0.25, 0.25, 1}

\definecolor{thirdbestcolor}{rgb}{1,1, 0.6}
\definecolor{secondbestcolor}{rgb}{1, 0.9, 0.6}
\definecolor{firstbestcolor}{rgb}{1, 0.6, 0.6}

\definecolor{m_green}{rgb}{0.57, 0.69, 0.41}
\definecolor{m_blue}{rgb}{0.47, 0.56, 0.74}
\definecolor{m_red}{rgb}{0.64, 0.35, 0.31}
\definecolor{l_green}{rgb}{0.85,0.90,0.83}
\definecolor{l_blue}{rgb}{0.87,0.91,0.98}
\definecolor{l_red}{rgb}{0.94,0.82,00.80}

\usepackage{enumitem}
\setlist[itemize]{noitemsep,leftmargin=.15in,topsep=.2em}
\setlist[enumerate]{noitemsep,leftmargin=*,topsep=.2em}

\newcommand{\colorsquare}[1]{{\color{#1}$\blacksquare$}}

\newcommand{\newcolorsquare}[2]{{\color{#1}$\blacksquare$}\hspace{-7pt}\textcolor{#2}{\textbf{$\square$}}}

\newcommand{\name}{Segment3D}

\newcommand{\bh}{\mathbf{h}}

\newcommand{\bK}{\mathbf{K}}

\newcommand{\bp}{\mathbf{p}}\newcommand{\bP}{\mathbf{P}}

\newcommand{\bR}{\mathbf{R}}

\newcommand{\bt}{\mathbf{t}}\newcommand{\bT}{\mathbf{T}}

\newcommand{\nR}{\mathbb{R}}

\newcommand{\cL}{\mathcal{L}}

\newcommand{\figref}[1]{Fig.~\ref{#1}}
\newcommand{\secref}[1]{Sec.~\ref{#1}}

\newcommand{\eqnref}[1]{Eq.~\eqref{#1}}
\newcommand{\tabref}[1]{Table~\ref{#1}}

\makeatletter
\DeclareRobustCommand\onedot{\futurelet\@let@token\@onedot}
\def\@onedot{\ifx\@let@token.\else.\null\fi\xspace}
\def\eg{e.g\onedot} 
\def\ie{i.e\onedot} 
 
\def\etc{etc\onedot}

\def\etal{et~al\onedot}

\makeatother

\newcommand{\boldparagraph}[1]{\vspace{0.5em}\noindent{\bf #1.}}

\renewcommand{\paragraph}[1]{\boldparagraph{#1}}

\definecolor{darkgreen}{rgb}{0,0.7,0}
\definecolor{newyellow}{rgb}{1,0.8,0.05}
\definecolor{newgreen}{rgb}{0.2,0.8,0.2}

\newcommand{\xmark}{\ding{55}}%

\newcolumntype{P}[1]{>{\centering\arraybackslash}m{#1}}

\begin{document}

\title{\vspace{-25px}Segment3D: Learning Fine-Grained Class-Agnostic 3D Segmentation\\without Manual Labels}

\renewcommand*{\thefootnote}{\fnsymbol{footnote}}

\twocolumn[{%
\renewcommand\twocolumn[1][]{#1}%
\maketitle
\begin{center}
\captionsetup{type=figure}
\vspace{-30px}
\begin{overpic}[width=\textwidth,abs,unit=1mm,scale=.25,]{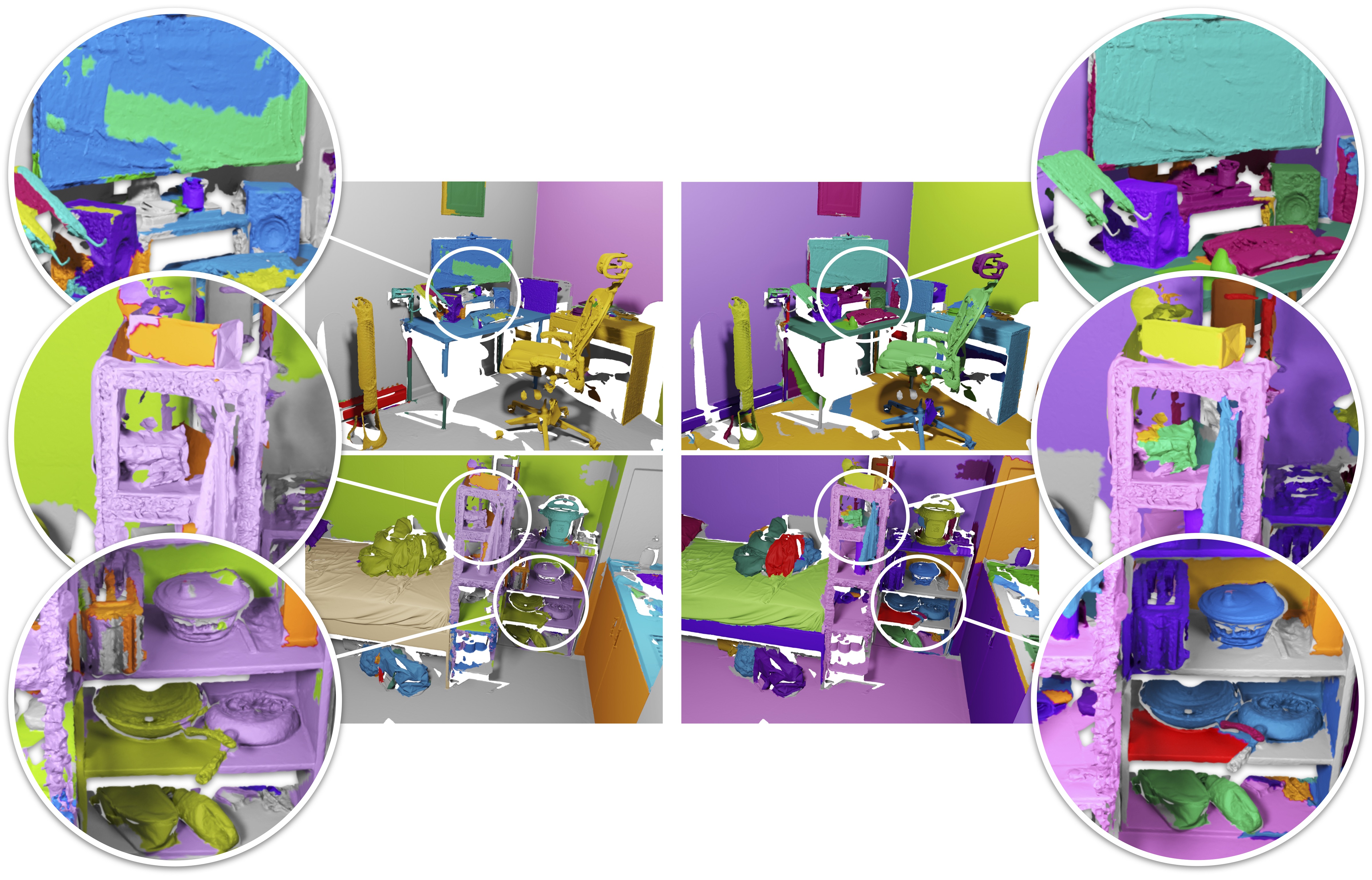}

\put(12,114){\large 
Rui Huang$^{1}$ \quad
Songyou Peng$^{2}$ \quad
Ayça Takmaz$^{2}$ \quad
Federico Tombari$^{3}$ \quad
Marc Pollefeys$^{2,4}$ }
\put(46,108){\large 
Shiji Song$^{1}$  \quad
Gao Huang$^{1}$${{\hypersetup{linkcolor=black}\footnotemark}}$ \quad
Francis Engelmann$^{2,3}$}

\put(44,100){ 
$^{1}$Tsinghua University \hspace{1mm}
$^{2}$ETH Zurich   \hspace{1mm}
$^{3}$Google   \hspace{1mm}
$^{4}$Microsoft}

\put(66,94){ \href{http://segment3d.github.io}{https://segment3d.github.io}}

\put(42,12){\parbox{4cm}{\center \it Mask3D\cite{Schult2023ICRA}\\trained on \textbf{manual} labels.}}
\put(91,13.8){\parbox{4.5cm}{\center \it Segment3D (Ours)\\trained on \textbf{automatic} labels.}}
\end{overpic}

\captionof{figure}{
\label{fig:teaser}
\textbf{Fine-Grained Class-Agnostic 3D Point Cloud Segmentation.}
\name{} predicts highly accurate segmentation masks \emph{(right)}, 
improves over state-of-the-art 3D segmentation methods (\eg{}, Mask3D~\cite{Schult2023ICRA}, \emph{left}), and does not require manually labeled 3D training data.
This is achieved through the automatic generation of high-quality training masks using foundation models for image segmentation\,\cite{Kirillov2023ICCV}.
}
\end{center}
}]
\footnotetext{Corresponding author.}

\begin{abstract}
\vspace{-5px}

Current 3D scene segmentation methods are heavily dependent on manually annotated 3D training datasets.
Such manual annotations are labor-intensive, and often lack fine-grained details.
Importantly, models trained on this data typically struggle to recognize object classes beyond the annotated classes, \ie{}, they do not generalize well to unseen domains and require additional domain-specific annotations.
In contrast, 2D foundation models demonstrate strong generalization and impressive zero-shot abilities, 
inspiring us to incorporate these characteristics from 2D models into 3D models.
Therefore, we explore the use of image segmentation foundation models to automatically generate training labels for 3D segmentation.
We propose \name{}, a method for class-agnostic 3D scene segmentation that produces high-quality 3D segmentation masks.
It improves over existing 3D segmentation models (especially on fine-grained masks), and enables easily adding new training data to further boost the segmentation performance -- all without the need for manual training labels.
\end{abstract}

\vspace{-15px}
\section{Introduction}~\label{sec:intro}
\vspace{-15px}

In this work, we propose \name{}, a method for fine-grained class-agnostic 3D segmentation.
In particular, dividing the space into coherent segments aligned with both the scene geometry and its semantics is a key challenge.
This ability to accurately segment and interpret 3D scenes is fundamental for intelligent assistants, such as autonomous robots or AR/VR devices that can help vision-impaired people navigate and engage with unknown spaces.

Current methods for 3D indoor-scene understanding mostly focus on semantic \cite{Nekrasov2021THREEDV,Schult2020CVPR, Qi2017CVPR, Qi2017NIPS, thomas2019kpconv,weder2023alster} and instance segmentation \cite{Schult2023ICRA, Lu2023ICCV, Chen2021ICCV,Liang2021CVPR,Vu2022CVPR,engelmann20203d}.
These approaches, while effective in benchmarking, have limitations stemming from their training.
Primarily, they depend on extensive manually labeled 3D training sets that are both time-consuming and challenging to annotate.
Importantly, their performance often deteriorates when applied to scenarios beyond their training data, limiting their effectiveness in diverse, real-world scenarios.
This becomes particularly apparent under the recently emerging task of \emph{open-vocabulary} 3D scene understanding \cite{Peng2023CVPR, Takmaz2023NIPS, Kerr2023ICCV, Lu2023CVPR} that aims to segment arbitrary user queries, which naturally go beyond the pre-defined set of training-set classes.
Concurrently, the recent surge in foundation models, particularly 2D vision-language models~\cite{Radford2021ICML,Jia2021ICML,Kirillov2023ICCV}, demonstrates remarkable potential.
Trained on internet-scale data, these models exhibit an extraordinary ability to generalize, even in a zero-shot setting, to new and different input distributions.
However, their application has been predominantly confined to 2D data. 
For instance, SAM~\cite{Kirillov2023ICCV} has shown impressive results in 2D image segmentation, but its applicability to 3D scene understanding remains mostly unexplored.
All these factors give rise to the interesting and significant research question:
\begin{center}
    \emph{How to leverage 2D foundation models for class-agnostic 3D scene segmentation without manually labeled 3D data?}
\end{center}

Recently, SAM3D~\cite{Yang2023ARXIV} has proposed a straightforward method that uses posed RGB-D images corresponding to a 3D scene. They first predict segmentation masks for RGB-D images with SAM~\cite{Kirillov2023ICCV}.
Subsequently, these 2D masks are projected into the 3D space, and through an iterative bottom-up process, the 3D masks of partial scenes are merged to derive the final segmentation result for the entire scene (see~\figref{fig:vssam3d}).
However, due to the heuristic merging rules, there are cases where masks that should be merged are not actually merged when combining the 3D masks of two adjacent frames. Additionally, variations in perspectives across frames can lead to conflicting segmentations in overlapping regions, which introduces noise during the merging.
Meanwhile, the inference speed of SAM3D is significantly slow, attributed to the application of SAM to a large number of RGB images in the scenes coupled with the bottom-up merging approach.

To circumvent the cumbersome and noisy merging procedure, we introduce Segment3D, a \emph{clean and efficient} approach that directly utilizes a \emph{native} 3D segmentation model to achieve class-agnostic, fine-grained 3D segmentation.
Segment3D employs a two-stage training approach that requires no hand-annotated labels for supervision.
We first pre-train our class-agnostic 3D segmentation model with the automatically generated 2D masks from SAM which are projected to partial RGB-D point clouds. 
Since there is readily available large-scale RGB-D data, additional data can be effortlessly integrated to further enhance segmentation performance.
This pre-training stage lays the groundwork for understanding the 3D structure from 2D annotations. However, as our ultimate objective is the segmentation of full 3D scenes, we must bridge the domain gap between partial point clouds and the more comprehensive 3D point clouds obtained from 3D scanners or reconstruction techniques~\cite{dai2017bundlefusion,izadi2011kinectfusion}.
To this end, in the second stage, we fine-tune the model on full 3D point clouds in a self-supervised manner, utilizing high-confidence mask predictions from the pre-trained model as training signal.

In experiments, we show strong performance in class-agnostic segmentation on ScanNet++~\cite{Yeshwanth2023ICCV} and further show the use of \name{} for improving open-vocabulary 3D instance segmentation ~\cite{Takmaz2023NIPS}.

Overall, the contributions of this paper are as follows:
\begin{itemize}
    \item We introduce Segment3D, a novel approach and training strategy for fine-grained class-agnostic 3D point cloud segmentation without manually annotated labels.
    \item Improved segmentation performance compared to a wide range of baselines, including fully supervised methods trained on carefully annotated datasets.
    \item We show that utilizing 2D foundation models to automatically generate high-quality training masks is a viable alternative to costly manual labeling.
\end{itemize}
\section{Related Work}~\label{sec:related}

\vspace{-20px}
\paragraph{3D Instance Segmentation}
Current models have seen a significant development over the last years,
from proposal-based~\cite{Yi2019CVPR,Yang2019NIPS}, over
grouping-based~\cite{Jiang2020CVPR, Chen2021ICCV, Liang2021CVPR, Vu2022CVPR,kreuzberg2022stop},
to recent Transformer-based~\cite{Schult2023ICRA,Lu2023ICCV} methods.
In this work, we follow the currently best-performing Transformer-based paradigm.
Despite impressive advancements, a shared limitation is the dependence on costly manual ground-truth annotations.
Recently, there have been efforts to automate the annotation process by bundling state-of-the-art segmentation models \cite{Weder2024labelmaker}, but they are still limited to pre-defined object classes.
However, in the context of open-world 3D scene understanding, the importance of semantic classification for closed-set categories has diminished.
A rare exception are interactive 3D segmentation methods\cite{kontogianni2023interactive, yue2023agile3d} where a human iteratively collaborates with a deep learning model to segment arbitrary objects, but naturally requires user input. Instead, in this work, we propose a general 3D segmentation method trained on automatically generated labels.

\paragraph{Foundation Models}
Foundation models, particularly those that are multimodal~\cite{Radford2021ICML,Jia2021ICML}, have revolutionized the field of AI by leveraging extensive image-text pre-training.
These models can derive rich image representations guided by natural language descriptions, enabling a variety of downstream tasks~\cite{Gu2022ICLR,Rao2021CVPR,Patashnik2021ICCV}. 
Another line of work~\cite{Caron2021ICCV,Oquab2023ARXIV}, based on self-supervised learning, employs image training and yields high-performance features directly applicable as inputs for linear classifiers.
Segment Anything Model (SAM)~\cite{Kirillov2023ICCV} has recently advanced the performance of foundation models for image segmentation.
SAM has undergone training on a diverse, high-quality dataset comprising more than 1 billion masks.
This training equips SAM with the ability to generalize to novel object types and images, surpassing the scope of its observations during the training process.
Additionally, SAM can generate high-quality, fine-grained masks. 
Our work leverages the power of SAM for pre-training a 3D segmentation model, which is later also used for generating supervision signals for fine-tuning on scene-level point clouds.

\boldparagraph{Open-Vocabulary 3D Scene Understanding}
Recently, there has been an increased interest in 3D open-vocabulary scene understanding.
This new field utilizes the zero-shot recognition abilities of 2D vision-language models~\cite{Radford2021ICML,Jia2021ICML}, enabling a more comprehensive understanding of diverse and previously unseen 3D environments~\cite{Zhang2022CVPR,Huang2023ICCV,Ha2022CORL,Peng2023CVPR,Kerr2023ICCV,Kobayashi2022NIPS,Ding2023CVPR,Lu2023CVPR,Zeng2023CVPR,Takmaz2023NIPS,Chen2023CVPR}.
PLA~\cite{Ding2023CVPR} aligns point cloud features with captions extracted from multi-view images of a scene to enable open-vocabulary recognition. 
OpenScene~\cite{Peng2023CVPR} distills per-pixel image features to 3D point clouds,
generating point-wise scene representations co-embedded with text and image pixels in CLIP feature space.
However, it mainly focuses on semantic segmentation and exhibits a limited understanding of object instances. 
To this end, OpenMask3D~\cite{Takmaz2023NIPS} predicts class-agnostic 3D instance masks and aggregates per-mask features via multi-view fusion of CLIP-based image embeddings.
Although they have achieved impressive results in open-vocabulary tasks,
a highly versatile segmentation model is required to accomplish this.
Our method contributes to this area by providing a variety of class-agnostic masks.
These masks can serve as foundational inputs for models like OpenMask3D,
enhancing their ability to perform instance-related tasks.

\section{Method}
\label{sec:method}
\begin{figure}
\centering
\includegraphics[width=0.46\textwidth]{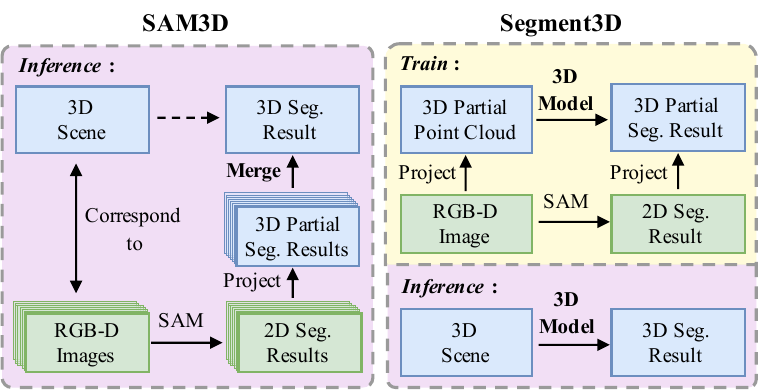}
\caption{\textbf{SAM3D vs. Segment3D.}
SAM3D~\cite{Yang2023ARXIV} \emph{(left)} merges 2D segmentation masks of RGB-D images generated by SAM to obtain 3D segmentation of entire scenes. 
This merging process introduces noise due to heuristic merging rules and conflict segmentation results across overlapping frames. Moreover, it is slow because of extensive image inference and the cumbersome merging procedure.
Instead, our Segment3D \emph{(right)} utilizes a native 3D model to directly segment entire 3D scenes. %
}
\vspace{-8pt}
\label{fig:vssam3d}
\end{figure}

\begin{figure*}[ht]
    \centering
    \includegraphics[width=0.99\textwidth]{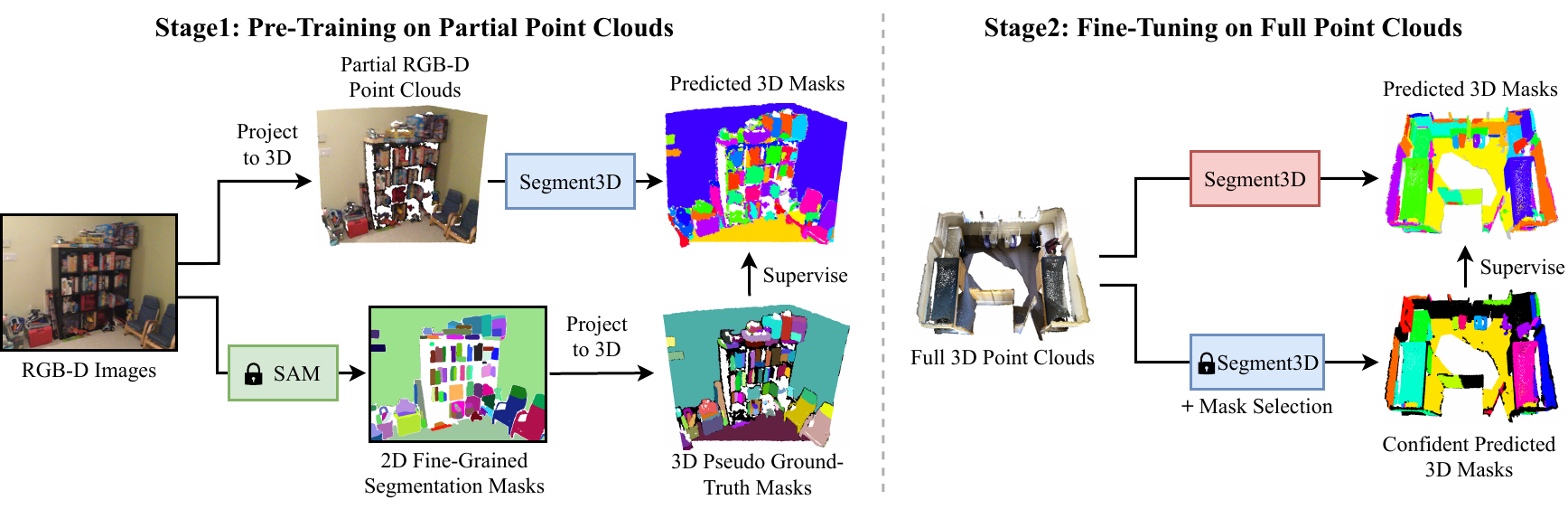}
    \caption{\textbf{Method Overview.}
    Training \name{} involves two stages:
    The first stage \emph{(left)} relies on largely available RGB-D image sequences and SAM \newcolorsquare{l_green}{m_green}, a pre-trained foundation model for 2D image segmentation \cite{Kirillov2023ICCV}.
    \name{} \newcolorsquare{l_blue}{m_blue} is pre-trained on partial RGB-D point clouds and supervised with pseudo ground-truth masks from SAM projected to 3D. Due to the domain gap between partial and full point clouds, in the second stage \emph{(right)}, \name{} \newcolorsquare{l_red}{m_red} is fine-tuned with confident masks predicted by the pre-trained \name{} \newcolorsquare{l_blue}{m_blue}.
    }
    \vspace{-10pt}
    \label{fig:pipeline}
\end{figure*}
We are interested in a method that can segment any object in a given 3D scene.
Hence, depending on existing 3D training datasets cannot accomplish this goal,
as their mask annotations are limited to a predefined set of object classes \cite{Dai2017CVPR, Baruch2021ARXIV, straub2019replica, Armeni2016} and models trained on these datasets might not generalize well to novel classes. SAM\,\cite{Kirillov2023ICCV} is a foundation model for 2D image segmentation, which shows extraordinary generalization ability.
SAM3D~\cite{Yang2023ARXIV} introduced a straightforward method utilizing posed RGB-D images for 3D scene understanding. As shown in \figref{fig:vssam3d}, it employs SAM to predict segmentation masks for RGB-D images, followed by projecting these 2D masks into 3D space. Through an iterative bottom-up process, 3D masks of partial scenes are merged to obtain the segmentation result for the entire scene. However, heuristic merging rules sometimes hinder the correct merging of 3D masks, and conflicting segmentation results across overlapping frames can introduce noise during the merging.
Furthermore, the extensive image processing and the cumbersome merging procedure lead to the slow inference speed.
To avoid the cumbersome and noisy merging process, we introduce Segment3D, a clean and efficient approach which utilizes a native 3D model to directly segment the entire scene.

Our key idea is to automatically generate class-agnostic mask annotations using pre-trained foundation models \cite{bommasani2021opportunities}.
To employ SAM for our 3D segmentation model we formulate a two-stage training approach as illustrated in Fig.\,\ref{fig:pipeline}.
First, 2D masks generated with SAM are projected onto \emph{partial} RGB-D point clouds and serve as supervision signal for pre-training our class-agnostic 3D segmentation model (\secref{train_on_paritial}).
Since the final goal is to segment 3D scenes, we need to bridge the inevitable domain gap between partial RGB-D point clouds and full 3D point clouds from 3D scanners or reconstruction methods.
We therefore fine-tune our model on \emph{full} 3D point clouds using high-confident mask predictions from our pre-trained model (\secref{train_on_complete}).

\subsection{Stage 1: Pre-Training on RGB-D Point Clouds}
\label{train_on_paritial}
\vspace{-5pt}
In contrast to the relatively scarce availability of 3D data, there is an abundance of 2D data, particularly RGB-D images, which are readily accessible.
For example, ScanNet~\cite{Dai2017CVPR} comprises merely 1513 3D scans, compared to the substantially larger collection of  $2.5$ million RGB-D images.
Leveraging the automatically generated masks from SAM together with the abundant 2D data,
we pre-train our 3D segmentation model on partial RGB-D point clouds.

\boldparagraph{Data Preparation}
Starting from a collection of RGB-D frames,
we create partial 3D point clouds and their corresponding pseudo ground-truth 3D masks.
Note that the labels can be automatically obtained without manual effort.

For each frame in a large RGB-D dataset, we first transform the 2D depth map to a partial 3D point cloud. To do so, we need to know the intrinsic matrix $\bK\in\nR^{3\times 3}$ and extrinsic matrix $\bT = \begin{bmatrix} \bR &\bt\end{bmatrix}\in\nR^{3\times 4}$.
For each pixel $\bp =(u, v)$, we can transform it with its depth value $D_\bp$ into a 3D point $\bP$ in world coordinates as follows:
\vspace{-5pt}
\begin{equation}
\bP = \bR^\top \cdot(D_\bp\cdot \bK^{-1}\cdot \tilde{\bp}) - \bR^\top \bt
\label{eq:unproject}
\vspace{-5pt}
\end{equation}
where $\tilde{\bp}$ is the homogeneous coordinate of $\bp$.
By applying the process in~\eqnref{eq:unproject} to all pixels in the depth map $D$ and associating their per-pixel RGB value, we obtain the input partial 3D point cloud.
Now, to obtain the {pseudo ground-truth} 3D segmentation masks for this point cloud, we prompt SAM~\cite{Kirillov2023ICCV} with a regular grid on the RGB image and acquire on average {$\sim50$} high-quality 2D segmentation masks per frame.
Since we know the one-to-one mapping between 2D pixel and 3D points from ~\eqnref{eq:unproject}, we directly obtain the per-point 3D mask labels.

\boldparagraph{Model Architecture}
We use a model inspired by Mask3D~\cite{Schult2023ICRA} to train a class-agnostic 3D segmentation model.
The model is comprised of a sparse convolutional backbone derived from MinkowskiUNet~\cite{Choy2019CVPR} and a transformer decoder, as in MaskFormer~\cite{cheng2021per, cheng2022masked}.
We adopt a set of queries to represent the masks,
each of which is initialized with a positional embedding.
Specifically, we select query positions with furthest point sampling (FPS) and use their Fourier positional encodings as the query embeddings.
Leveraging the transformer decoder,
all the mask queries are refined by progressively attending to point cloud features across multiple scales in parallel.
Each mask query is subsequently decoded into both a mask feature and a binary label to predict whether the given query corresponds to a valid object or not.
By computing cosine similarity scores between a mask feature and all point features within the point cloud, a heatmap is generated over the point cloud.
This heatmap is input to a sigmoid function, and thresholded at 0.5, resulting in the final binary mask.

\boldparagraph{Training with SAM Generated Masks}
We supervise the model with two losses:
the per-point level supervision loss $\mathcal{L}_{\text {mask}}$ and a {per-query} level supervision loss $\mathcal{L}_{\text {obj}}$.
The loss $\mathcal{L}_\text {mask}$ enables learning a foreground-background segmentation for each mask, and is composed of a dice loss $\cL_\text{dice}$~\cite{Milletari20163DV} and a binary cross-entropy loss $\cL_{\mathrm{ce}}$ for each point.
The $\mathcal{L}_{\mathrm{obj}}$ is a binary classification loss that indicates whether a query represents a valid ``object" or ``no object".
This mechanism allows for the prediction of a variable number of masks, depending on the underlying scene content and geometry.
Following prior work~\cite{carion2020end,Schult2023ICRA,cheng2022masked}, we first adopt bipartite graph matching to establish correspondences between the set of predicted masks and the set of target masks originating from SAM as described before.
If the predicted instance finds a matching target mask, then we assign it an ``object" label; conversely, if there is no match, we assign ``no object".
In summary, we optimize the following losses:
\vspace{-5pt}
\begin{gather}
\mathcal{L}=\mathcal{L}_{\text {mask}}+\lambda_{\text{obj}}\mathcal{L}_{\text{obj}} \\
\mathcal{L}_{\text {mask}}=\lambda_{\text {dice}} \mathcal{L}_{\text {dice}}+\lambda_{\mathrm{ce}} \mathcal{L}_{\mathrm{ce}}
\label{eq:total_loss}
\end{gather}
where $\lambda_{*}$ are hyperparameters that balance the contribution of each component in the loss. 
The binary classification loss $\mathcal{L}_{\text{obj}}$ is applied to all queries, while the mask loss $\mathcal{L}{_\text{mask}}$ is specifically applied to masks labeled as ``object".

\subsection{Stage 2: Self-Supervised Scene Fine-Tuning}
\label{train_on_complete}
\begin{figure}
\centering
\includegraphics[width=0.47\textwidth]{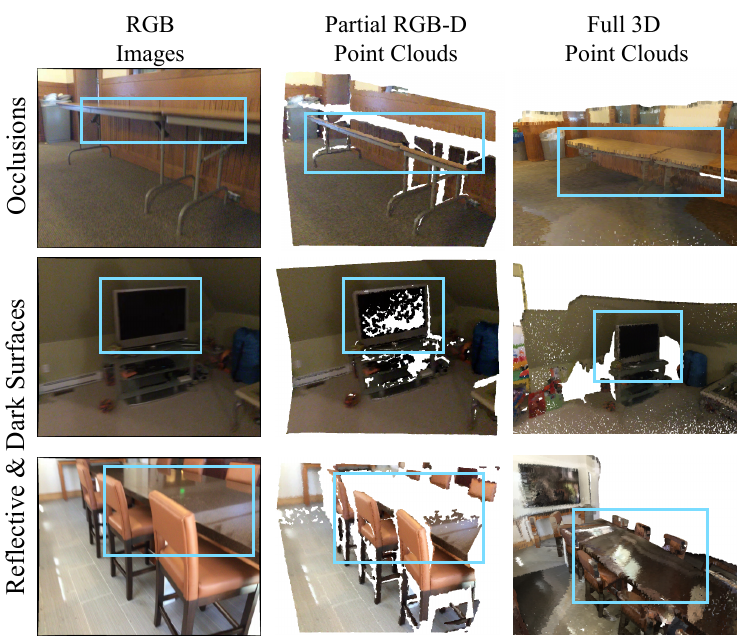}
\caption{\textbf{Partial vs. Full Point Clouds.}
We show the difference between partial \emph{(center)} and full \emph{(right)} point clouds from ScanNet~\cite{Dai2017CVPR}.
Full point clouds are more complete and exhibit fewer occlusions due to reconstruction over multiple viewpoints. 
Due to this domain gap, fine-tuning on full 3D point clouds is necessary.
}
\vspace{-10pt}
\label{fig:domain_gap}
\end{figure}

After {Stage 1}, we obtain a class-agnostic 3D segmentation model by pre-training solely on RGB-D images and automatically generated labels from SAM.
However, a fundamental domain gap persists between partial point clouds derived from RGB-D images and full point clouds acquired through a 3D scanner or reconstruction methods \cite{izadi2011kinectfusion,dai2017bundlefusion} (See Fig.\,\ref{fig:domain_gap}).
This gap exists mostly because of object occlusions, but also due to challenges of depth cameras to capture dark or reflective surfaces from a single viewpoint.
Hence, depending solely on RGB-D frames for training a 3D segmentation model intended for full 3D scans proves inadequate.
Therefore, we propose to further fine-tune our model on scene-level full 3D point clouds.
The key idea to obtain 3D mask annotations for training on full point clouds is to use selected, high-confidence masks generated by the pre-trained model itself.
Note that this approach does not require any manual labels on full 3D scenes and proves essential for the performance of \name{} (see Tab.\,\ref{tab:ablate_stages}).

\newcommand{\tableclassagnostic}{
\begin{table*}[ht]
    \centering
    \small
    \setlength{\tabcolsep}{0.3cm}
    \resizebox{0.98\textwidth}{!}{
        
\begin{tabular}{lc ccc ccc}
\toprule
& & \multicolumn{3}{c}{\textit{without post-processing}} & \multicolumn{3}{c}{\textit{with post-processing}}\\
\cmidrule(lr){3-5} \cmidrule(lr){6-8}
\textbf{Model} & \textbf{Ground Truth Labels} & \textbf{AP} & \textbf{AP$_{50}$} & \textbf{AP$_{25}$}  & \textbf{AP} & \textbf{AP$_{50}$} & \textbf{AP$_{25}$} \\
\midrule
SAM3D~\cite{Yang2023ARXIV} & \xmark & $3.9$ & $9.3$ & $22.1$ & $8.4$ & $16.1$ & $30.0$ \\ 
Felzenszwalb \emph{et al.}~\cite{Felzenszwalb2004IJCV} & \xmark & $5.8$ & $11.6$ & $27.2$ & -- & -- & -- \\
Mask3D~\cite{Schult2023ICRA} & ScanNet200~\cite{Rozenberszki2022ARXIV} & $8.7$ & $15.5$ & $27.2$ & $14.3$ & $21.3$ & $29.9$ \\
Mask3D~\cite{Schult2023ICRA} & ScanNet~\cite{Dai2017CVPR} & 9.4 & 16.8 & 28.7 & 15.4 & 22.7 & 31.6 \\
\midrule
\name{} (\textbf{Ours}) & \xmark &
$\mathbf{12.0}$ &
$\mathbf{22.7}$ &
$\mathbf{37.8}$ &
$\mathbf{19.0}$ &
$\mathbf{29.7}$ &
$\mathbf{41.6}$ \\
& &
\textcolor{darkgreen}{\footnotesize{(+\,$2.6$)}} &
\textcolor{darkgreen}{\footnotesize{(+\,$5.9$)}} &
\textcolor{darkgreen}{\footnotesize{(+\,$9.1$)}} &
\textcolor{darkgreen}{\footnotesize{(+\,$3.6$)}} &
\textcolor{darkgreen}{\footnotesize{(+\,$7.0$)}} &
\textcolor{darkgreen}{\footnotesize{(+\,$10.0$)}}
\\
\bottomrule
\end{tabular}

        }
        \vspace{3px}
    \caption{
        \textbf{3D Segmentation Scores on ScanNet++ Val Set.}
        The evaluation metric is average precision (AP).
        Similar to \cite{Felzenszwalb2004IJCV, Yang2023ARXIV}, \name{} does not require manually annotated training labels. We report scores with and without post-processing (more details in Sec.~\ref{sec:sota}).
        }
        \vspace{-15pt}
\label{tab:class_agnostic}
\end{table*}
}

\newcommand{\tableobjectsize}{

\begin{table}[b]
    \centering
    \small
    \vspace{-5pt}
\begin{tabular}{cc cc}
\toprule
\textbf{Mask Size} & & \textbf{Mask3D}\cite{Schult2023ICRA} & \textbf{Segment3D} (Ours)\\
\midrule
\multirow{2}{*}{Large} & AP$_{50}$ & $30.6$ & $30.4$ \textcolor{orange}{\footnotesize{(-\,$0.2$)}} \\
                       & AP$_{25}$ & $47.7$ & $47.0$ \textcolor{orange}{\footnotesize{(-\,$0.7$)}} \\
\midrule
\multirow{2}{*}{Medium}& AP$_{50}$ & $25.8$ & $29.2$ \textcolor{darkgreen}{\footnotesize{(+\,$3.4$)}} \\
                       & AP$_{25}$ & $41.8$ & $49.5$ \textcolor{darkgreen}{\footnotesize{(+\,$7.7$)}} \\
\midrule
\multirow{2}{*}{Small} & AP$_{50}$ & $4.5$ & $15.9$ \textcolor{darkgreen}{\footnotesize{(+\,$11.4$)}} \\
                       & AP$_{25}$ & $11.3$ & $29.6$ \textcolor{darkgreen}{\footnotesize{(+\,$18.3$)}} \\
\bottomrule
\end{tabular}
\vspace{3px}
\caption{
\textbf{Segmentation Scores on Different Mask Sizes.}
The evaluation metric is average precision (AP) on ScanNet++ validation set.
Segmented3D improves over Mask3D, especially on small and medium-sized object masks. Details are in Sec.\,\ref{sec:analysis}.
}
\label{tab:object_size}
\end{table}

}

\newcommand{\tableablatestages}{
\begin{table}[h]
    \centering
    \small
    \setlength{\tabcolsep}{0.1cm}
    \resizebox{0.48\textwidth}{!}{
        \begin{tabular}{l|lll}
\toprule
     \textbf{Training Stages} & \textbf{AP} & \textbf{AP$_{50}$} & \textbf{AP$_{25}$} \\
     \midrule
     Pre-Training (Stage~1) & $7.4$ & $15.2$ & $31.2$ \\
     + Fine-Tuning (Stage~2)  &
     $\mathbf{12.0}$ \textcolor{darkgreen}{\footnotesize{(+\,$4.6$)}} &
     $\mathbf{22.7}$ \textcolor{darkgreen}{\footnotesize{(+\,$7.5$)}} &
     $\mathbf{37.8}$ \textcolor{darkgreen}{\footnotesize{(+\,$6.6$)}} \\
     \bottomrule
\end{tabular}

}
    \vspace{3px}
\caption{
        \textbf{Effect of Two-Stage Training.}
        Fine-tuning on full point clouds supervised by confident predicted 3D masks significantly surpasses pre-training on projected 2D SAM masks alone.}
    \vspace{-10pt}
\label{tab:ablate_stages}
\end{table}
}

\newcommand{\tablenumqueries}{
\begin{table}[!ht]
    \centering
    \small
    \setlength{\tabcolsep}{0.3cm}
    \resizebox{0.4\textwidth}{!}{
        \begin{tabular}{c|ccc}
\toprule
     Num of queries & \textbf{AP} & \textbf{AP$_{50}$} & \textbf{AP$_{25}$} \\
     \midrule
     100 & 8.5 & 16.7 & 30.3 \\
     150 & 9.8 & 19.2 & 34.6 \\
     200 & 10.7 & 20.9 & 36.6 \\
     250 & 11.3 & 21.7 & 37.0\\
     300 & 11.6 & 22.1 & 37.6\\
     350 & 11.9 & 22.5 & 37.8\\
     400 & 12.0 & 22.7 & 37.8\\
     \bottomrule
\end{tabular}
}
    \caption{
        \textbf{Performance on different number of queries for our method.}
        }
\label{tab:num_queries}
\end{table}
}

\newcommand{\tablemoredata}{
\begin{table}[!h]
    \centering
    \small
    \setlength{\tabcolsep}{0.1cm}
    \resizebox{0.48\textwidth}{!}{
        
\begin{tabular}{l|lll}
\toprule
     \textbf{Pre-Training Dataset} (\# Frames) & \textbf{AP} & \textbf{AP$_{50}$}\\
     \midrule
     ScanNet ($76$k)             & $12.0$ & $22.7$\\ %
     ScanNet ($76$k), ScanNet++ ($34$k) &
     $13.7$ \textcolor{darkgreen}{\footnotesize{(+\,$1.7$)}} &
     $24.7$ \textcolor{darkgreen}{\footnotesize{(+\,$2.0$)}} \\
     \bottomrule
\end{tabular}
}
    \caption{
        \textbf{Performance Increase with Additional Training Data.}
        }
        \vspace{-6px}
\label{tab:more_data}
\end{table}
}

\newcommand{\tableopenmask}{
\begin{table}[!h]
    \centering
    \small
    \setlength{\tabcolsep}{0.1cm}
    \resizebox{0.48\textwidth}{!}{
        \begin{tabular}{l|lll}
\toprule
     \textbf{Model} & \textbf{Segmentor} & \textbf{ScanNet++} & \textbf{Replica}\\
     \midrule
     OpenMask3D~\cite{Takmaz2023NIPS} & Mask3D\cite{Schult2023ICRA} & $15.0$ & $18.0$\\
     OpenMask3D~\cite{Takmaz2023NIPS} & \name{} (Ours) &
     $17.7$ \textcolor{darkgreen}{\footnotesize{(+\,$2.7$)}} &
     $18.7$ \textcolor{darkgreen}{\footnotesize{(+\,$0.7$)}} \\
     \bottomrule
\end{tabular}

}
    \caption{
        \textbf{Open-Set 3D Scene Understanding Scores.}
        The evaluation metric is average precision with an IoU threshold of $50$\%.}
        \vspace{-5px}
\label{tab:openmask3d}
\end{table}
}
\boldparagraph{Confidence-Based Mask Generation} 
Next, we outline the process of generating the supervision signal for the fine-tuning stage.
The pre-trained model processes point clouds independently of their nature, be it partial or full.
Therefore, when presented with a full 3D point cloud, the pre-trained 3D model produces a set of masks,
each with a binary classification (valid or not) and a heatmap over all points,
just as in \secref{train_on_paritial}.
To assess the quality of the predicted masks,
we compute a confidence score
based on the confidence map $\sigma(\bh)$, where $\bh$ is the predicted heatmap and $\sigma$ is the sigmoid function.
We then compute the average confidence of those points for which $\sigma(\bh)>0.5$ as the confidence score of the predicted mask, denoted as $c_{\text {mask}}$.
We also consider the classification as a valid object and use the probability from the binary classification assigned to the ``object" category as the confidence score, denote as $c_{\text {obj}}$.
The final confidence score for each predicted mask is then the product of the two scores, $c = c_{\text{mask}}\cdot c_{\text{obj}}$.
For fine-tuning our 3D segmentation model in Stage\,2, we select the most confidently predicted masks above a threshold $\tau_c$.

\boldparagraph{Training with the High-Confidence Generated Masks}
For fine-tuning, we follow the same procedure as before and use $\mathcal{L}_\text{mask}$ as defined in Eq.\,\ref{eq:total_loss}.
In contrast to the pre-training stage, the binary classification loss $\mathcal{L}_\text{obj}$,
responsible for categorizing queries into valid or invalid, is omitted.
Since we only select masks with high confidence for supervision,
it can happen that some objects in the scene have no assigned ground truth mask.
In such instances, deeming a correctly predicted mask for those objects as invalid would be detrimental.
\tabref{tab:ablate_stages} illustrates the efficacy of the self-supervised fine-tuning process in comparison to pre-training alone.

\begin{table*}[ht]
    \centering
    \small
    \setlength{\tabcolsep}{0.3cm}
    \resizebox{0.98\textwidth}{!}{
        
\begin{tabular}{lc ccc ccc}
\toprule
& & \multicolumn{3}{c}{\textit{without post-processing}} & \multicolumn{3}{c}{\textit{with post-processing}}\\
\cmidrule(lr){3-5} \cmidrule(lr){6-8}
\textbf{Model} & \textbf{Ground Truth Labels} & \textbf{AP} & \textbf{AP$_{50}$} & \textbf{AP$_{25}$}  & \textbf{AP} & \textbf{AP$_{50}$} & \textbf{AP$_{25}$} \\
\midrule
SAM3D~\cite{Yang2023ARXIV} & \xmark & $3.9$ & $9.3$ & $22.1$ & $8.4$ & $16.1$ & $30.0$ \\ 
Felzenszwalb \emph{et al.}~\cite{Felzenszwalb2004IJCV} & \xmark & $5.8$ & $11.6$ & $27.2$ & -- & -- & -- \\
Mask3D~\cite{Schult2023ICRA} & ScanNet200~\cite{Rozenberszki2022ARXIV} & $8.7$ & $15.5$ & $27.2$ & $14.3$ & $21.3$ & $29.9$ \\
Mask3D~\cite{Schult2023ICRA} & ScanNet~\cite{Dai2017CVPR} & 9.4 & 16.8 & 28.7 & 15.4 & 22.7 & 31.6 \\
\midrule
\name{} (\textbf{Ours}) & \xmark &
$\mathbf{12.0}$ &
$\mathbf{22.7}$ &
$\mathbf{37.8}$ &
$\mathbf{19.0}$ &
$\mathbf{29.7}$ &
$\mathbf{41.6}$ \\
& &
\textcolor{darkgreen}{\footnotesize{(+\,$2.6$)}} &
\textcolor{darkgreen}{\footnotesize{(+\,$5.9$)}} &
\textcolor{darkgreen}{\footnotesize{(+\,$9.1$)}} &
\textcolor{darkgreen}{\footnotesize{(+\,$3.6$)}} &
\textcolor{darkgreen}{\footnotesize{(+\,$7.0$)}} &
\textcolor{darkgreen}{\footnotesize{(+\,$10.0$)}}
\\
\bottomrule
\end{tabular}

        }
        \vspace{3px}
    \caption{
        \textbf{3D Segmentation Scores on ScanNet++ Val Set.}
        The evaluation metric is average precision (AP).
        Similar to \cite{Felzenszwalb2004IJCV, Yang2023ARXIV}, \name{} does not require manually annotated training labels. We report scores with and without post-processing (more details in Sec.~\ref{sec:sota}).
        }
        \vspace{-15pt}
\label{tab:class_agnostic}
\end{table*}

\section{Experiments}~\label{sec:experiments}
We firstly compare with fully-supervised and traditional geometric segmentation baselines in a class-agnostic segmentation setting (Sec.\,\ref{sec:sota}).
We then provide detailed analysis to understand the importance of fine-tuning,
the effect of training on more data and show the advantage of our method in segmenting small objects (Sec.\,\ref{sec:analysis}).
We also show qualitative results of our 3D segmentation method (Sec.\,\ref{sec:quali}).
Finally, we demonstrate its potential application for the  task of open-set 3D instance segmentation as recently proposed in OpenMask3D~\cite{Takmaz2023NIPS} (Sec.\,\ref{sec:openset}).

\subsection{Comparing with State-of-the-Art Methods}
\label{sec:sota}
\vspace{-5pt}
\boldparagraph{Setup}
We run experiments on four popular datasets including ScanNet\,\cite{Dai2017CVPR}, its extension ScanNet200\,\cite{Rozenberszki2022ARXIV}, the Replica\cite{straub2019replica} dataset,
and the newly released ScanNet++\,\cite{Yeshwanth2023ICCV}.
For training our model, we employ ScanNet~\cite{Dai2017CVPR,Rozenberszki2022ARXIV},
which is collected through a lightweight RGB-D scanning process.
ScanNet comprises $1513$ indoor scenes,
encompassing $\sim$$2\cdot10^6$ views,
along with 3D camera poses,
surface reconstruction and instance-level semantic mask annotations.
For Stage 1, we sample every 25$^{\text{th}}$ frame of the RGB-D sequences ($\sim$~1\,FPS) and obtain approximately $76\cdot 10^3$ training frames.
For Stage 2, we use the $\sim$$1.2\cdot 10^3$ reconstructed 3D scans of indoor spaces as full point clouds.
For evaluation, we use ScanNet++~\cite{Yeshwanth2023ICCV} which comes with high-fidelity 3D mask annotations including smaller objects which are not well annotated in ScanNet. 
It includes high-resolution 3D scans captured at sub-millimeter precision and annotated comprehensively, covering objects of varying sizes. 
See the appendix for the implementation details of our method.

\boldparagraph{Methods in Comparison}
We compare with a wide range of prior art methods from different categories.
Mask3D\,\cite{Schult2023ICRA} is a fully-supervised transformer-based method trained on manually annotated 3D segmentation masks.
Transformer-based methods \cite{Schult2020CVPR, sun2023superpoint} currently define the state-of-the-art approach for 3D instance segmentation. \name{} has the same backbone as Mask3D but instead of training on manually annotated 3D masks, it learns from automatically generated 2D (pre-training) and 3D masks (fine-tuning).
Felzenszwalb \emph{\etal{}}~\cite{Felzenszwalb2004IJCV} proposed a graph-based method for segmentation which does not require any training data and operates directly on the 3D geometry.
As mentioned in \secref{sec:method}, SAM3D~\cite{Yang2023ARXIV} merges the segmentation masks of RGB-D images generated by SAM to obtain the 3D segmentation result. Similarly to ours, it does not rely on manual labels.

\boldparagraph{Metrics}
We evaluate all methods on the validation set of ScanNet++ 
and report average precision (AP) scores at IoU thresholds of $25$\%, $50$\%, and averaged over the range [0.5:0.95:05] between predicted and ground truth masks.
Consistent with common practices in the field \cite{Schult2020CVPR, sun2023superpoint, Vu2022CVPR}, we also report scores after post-processing.
This involves smoothing the predicted masks through graph-based oversegmentation \cite{Felzenszwalb2004IJCV}, and splitting distant parts of the same mask via connected component clustering DBSCAN \cite{ester1996density}. 

\boldparagraph{Results}
Scores are reported in \tabref{tab:class_agnostic}.
\name{} outperforms all previous methods by at least $\mathbf{+2.6}$ AP and up to $\mathbf{+10.0}$ AP$_{25}$.
Notably, we achieve such improvements \emph{without ground-truth mask annotations} as used by Mask3D.
In general, the performance of Mask3D (and other fully supervised methods) depends on the quality of the annotated training dataset; often the manual annotation of small objects (pens, cellphones) or other fine-details is challenging.
Instead, \name{} relies on automatically generated high-quality masks from SAM which can capture fine-grained details without human annotation effort.
Table~\ref{tab:object_size} highlights that \name{} excels particularly in predicting the more challenging small object masks.

Finally, our approach significantly outperforms SAM3D despite both methods relying on the same 2D masks from SAM.
Even without fine-tuning \name{} already outperforms SAM3D (see Table~\ref{tab:ablate_stages}). 
This shows that Segment3D can circumvent the noise that arises during the merging process of SAM3D. Furthermore, Segment3D is more efficient since it only processes a single input of point cloud, while SAM3D needs to inference a large number of RGB-D images and involves a bottom-up merging procedure.

{Note the comparable performance of Mask3D trained on ScanNet and ScanNet200, which we attribute to similar mask annotations in both datasets.
They differ only in the labeled classes (20 \emph{vs.} 200), while most categories of ScanNet200 are labeled as ``other" in ScanNet.
For both datasets, we utilize all available masks independent of their semantic class to maximize the number of training labels.}

\subsection{Analysis Experiments}
\label{sec:analysis}
\vspace{-5pt}

\begin{table}[b]
    \centering
    \small
    \vspace{-5pt}
\begin{tabular}{cc cc}
\toprule
\textbf{Mask Size} & & \textbf{Mask3D}\cite{Schult2023ICRA} & \textbf{Segment3D} (Ours)\\
\midrule
\multirow{2}{*}{Large} & AP$_{50}$ & $30.6$ & $30.4$ \textcolor{orange}{\footnotesize{(-\,$0.2$)}} \\
                       & AP$_{25}$ & $47.7$ & $47.0$ \textcolor{orange}{\footnotesize{(-\,$0.7$)}} \\
\midrule
\multirow{2}{*}{Medium}& AP$_{50}$ & $25.8$ & $29.2$ \textcolor{darkgreen}{\footnotesize{(+\,$3.4$)}} \\
                       & AP$_{25}$ & $41.8$ & $49.5$ \textcolor{darkgreen}{\footnotesize{(+\,$7.7$)}} \\
\midrule
\multirow{2}{*}{Small} & AP$_{50}$ & $4.5$ & $15.9$ \textcolor{darkgreen}{\footnotesize{(+\,$11.4$)}} \\
                       & AP$_{25}$ & $11.3$ & $29.6$ \textcolor{darkgreen}{\footnotesize{(+\,$18.3$)}} \\
\bottomrule
\end{tabular}
\vspace{3px}
\caption{
\textbf{Segmentation Scores on Different Mask Sizes.}
The evaluation metric is average precision (AP) on ScanNet++ validation set.
Segmented3D improves over Mask3D, especially on small and medium-sized object masks. Details are in Sec.\,\ref{sec:analysis}.
}
\label{tab:object_size}
\end{table}

\boldparagraph{Performance on Different Mask Sizes}
We proceed with an analysis of the performance of our \name{} and Mask3D (the best-performing baseline) on object masks of various sizes.
We categorize the size of masks based on the number of points they contain
(small: [$0$, $2$k],
medium: [$2$k, $15$k]
large: [$15$k, $\infty$])
and exclude the masks of the floor, ceiling and walls.
The results are reported in \tabref{tab:object_size}.
Our method yields significantly improved segmentation results on small and medium-sized masks, and performs on par with Mask3D on large masks.
This confirms our intuition that Mask3D performs poorly on small-sized object masks
as those are typically harder to manually annotate.
In contrast, \name{} utilizes masks from SAM as supervision, which capture fine-grained scene details.
In summary, our model, trained on automatically generated masks by a 2D foundation model, surpasses a model trained on manually labeled datasets.
This showcases the usefulness of foundation models and raises the question if manually labeled large-scale 3D datasets are necessary for training 3D point cloud segmentation models.

\begin{figure*}
    \centering
    \includegraphics[width=0.96\textwidth]{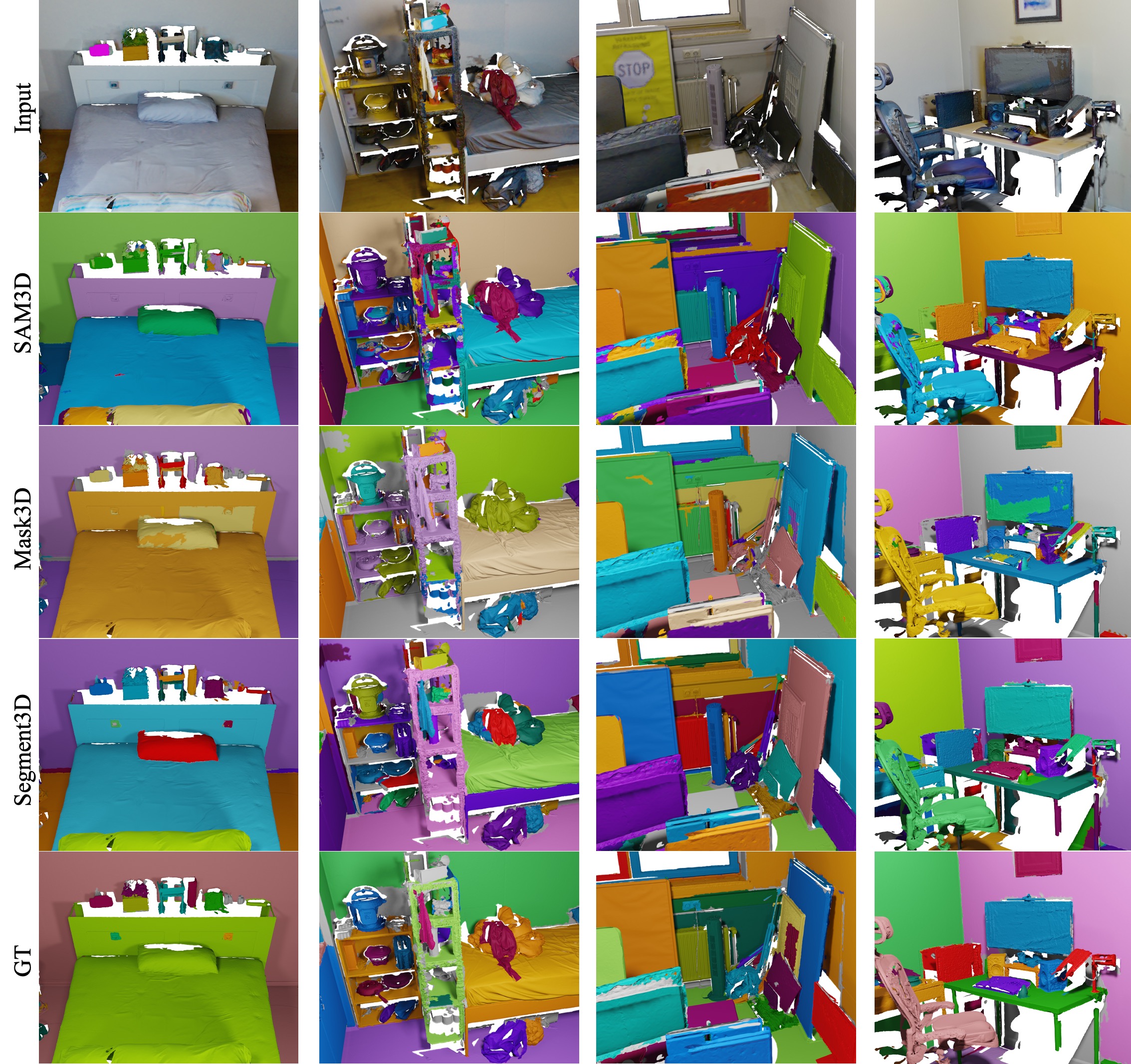}
    \vspace{-1px}
    \caption{\textbf{Qualitative Results on ScanNet++ Val Set.}
    From top to bottom, we show the colored input 3D scenes, the segmentation masks predicted by SAM3D~\cite{Yang2023ARXIV},
    Mask3D~\cite{Schult2023ICRA}, our \name{} and the ground truth 3D mask annotations.
    }
    \label{fig:quali}
\end{figure*}

\begin{figure*}
    \centering

\definecolor{amethyst}{rgb}{0.6, 0.4, 0.8}

\begin{overpic}[width=0.9\textwidth, ]{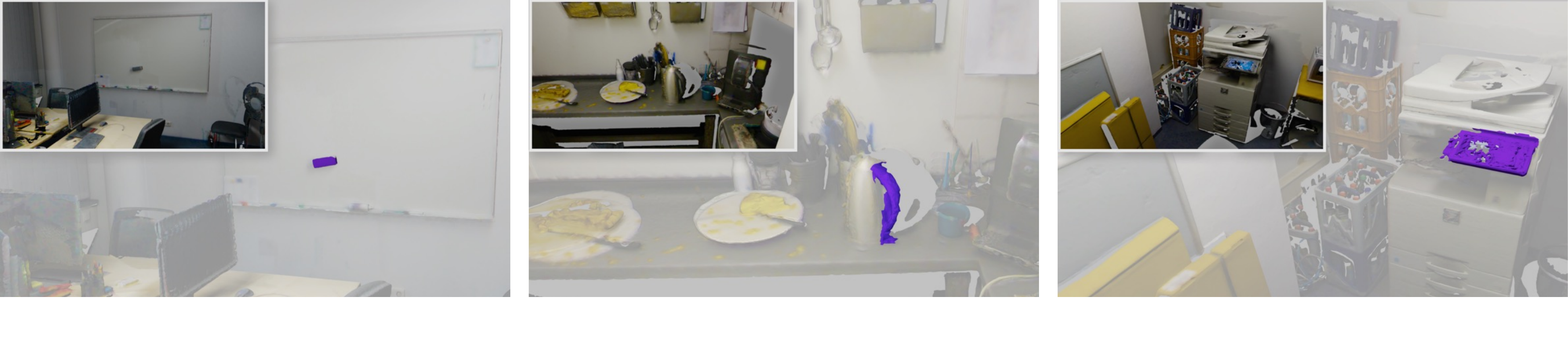}%
    \put(10,0.5){{\small \colorbox{white}
    {\textit{``a black eraser''}}}}
    \put(44,0.5){{\small   \colorbox{white}{\textit{``kettle handle"}}}}
    \put(74,0.5){{\small   \colorbox{white}{\textit{``copier control screen"}}}}
\end{overpic}
    \caption{\textbf{Qualitative Results of Adapted OpenMask3D~\cite{Takmaz2023NIPS}.}
    Given a text prompt \emph{(bottom)}, OpenMask3D finds the corresponding masks~\colorsquare{amethyst} in a given 3D scene \emph{(top)}. We show the 3D scene reconstruction and an RGB image for better visualization \emph{(top left corner)}.
    The original OpenMask3D based on Mask3D is unable to recognize any of the above queries, whether it is a relatively small object or a fine-grained affordance part of an object. In contrast, our \name{} adaptation performs well in these cases.
    }
    \label{fig:openmask3d}
\end{figure*}

\boldparagraph{The Effect of Two-Stage Training}
Next, we compare the performance of \name{} pre-trained solely on partial RGB-D point clouds (Stage~1)
and with additional fine-tuning on full point clouds (Stage~2). 
Scores are reported in \tabref{tab:ablate_stages}.
The additional fine-tuning stage almost doubles the segmentation performance of our model on the most challenging AP metric.
By training with the predicted high-confidence masks,
\name{} effectively reduces the inherent domain gap between the partial point clouds derived from RGB-D images and full 3D point clouds.

\begin{table}[h]
    \centering
    \small
    \setlength{\tabcolsep}{0.1cm}
    \resizebox{0.48\textwidth}{!}{
        \begin{tabular}{l|lll}
\toprule
     \textbf{Training Stages} & \textbf{AP} & \textbf{AP$_{50}$} & \textbf{AP$_{25}$} \\
     \midrule
     Pre-Training (Stage~1) & $7.4$ & $15.2$ & $31.2$ \\
     + Fine-Tuning (Stage~2)  &
     $\mathbf{12.0}$ \textcolor{darkgreen}{\footnotesize{(+\,$4.6$)}} &
     $\mathbf{22.7}$ \textcolor{darkgreen}{\footnotesize{(+\,$7.5$)}} &
     $\mathbf{37.8}$ \textcolor{darkgreen}{\footnotesize{(+\,$6.6$)}} \\
     \bottomrule
\end{tabular}

}
    \vspace{3px}
\caption{
        \textbf{Effect of Two-Stage Training.}
        Fine-tuning on full point clouds supervised by confident predicted 3D masks significantly surpasses pre-training on projected 2D SAM masks alone.}
    \vspace{-10pt}
\label{tab:ablate_stages}
\end{table}

\begin{figure}[h]
    \centering
    \definecolor{amber}{rgb}{1.0, 0.49, 0.0}
    \definecolor{azure}{rgb}{0.0, 0.5, 1.0}
     \hspace{5px}
    \colorsquare{darkgreen}{\footnotesize~Segment3D} \hspace{5px}
    \colorsquare{amber}{\footnotesize~Mask3D (ScanNet)} \hspace{5px}
    \colorsquare{azure}{\footnotesize~Mask3D (ScanNet200)}
    \\ \vspace{5px}
    \includegraphics[width=0.45\textwidth]{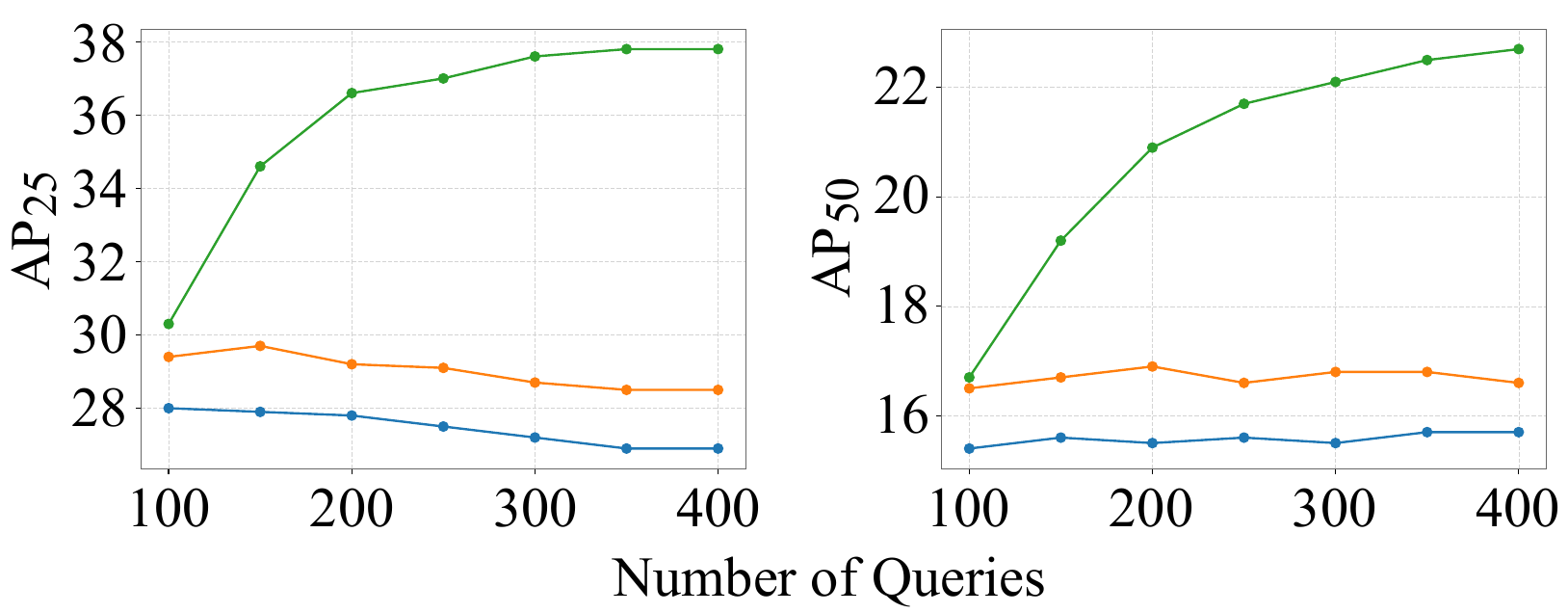}
    \caption{\textbf{Performance with Varying Number of Queries.} The metric is average precision with IoU thresholds of 25\% and 50\%.}
    \vspace{-10pt}
    \label{fig:num_queries}
\end{figure}

\boldparagraph{Number of Queries}
Since our model and Mask3D share the same transformer-based architecture,
we analyse the influence of the number of queries on the segmentation performance.
The results are presented in \figref{fig:num_queries}.
The number of queries is an important aspect of such models as each query represents a mask and ultimately defines the upper bound of recognizable masks in a 3D scene.
In the case of Mask3D, the segmentation performance remains relatively stable with changes in the number of queries, whereas \name{} experiences notable benefits from an increased number of queries.
This is an interesting observation since the key difference between Segment3D and Mask3D is the training data, and it shows the benefit of automatically generated SAM masks which are more diverse than the manually annotated ground truth masks in ScanNet.
Constrained by limited GPU memory, we were unable to assess performance with more than 400 queries. Nevertheless, the trend of the curves suggests the potential for further improved performance with increased query numbers.

\boldparagraph{Pre-Training with Additional Data}
Since RGB-D data is available in abundance and masks can be automatically generated,
it is natural to ask if pre-training on additional masks will further improve the overall performance.
To that end, we perform a first experiment where
we select additional frames from the training set of ScanNet++ dataset (to increase the variety of training data).
Again, we sample frames at roughly $1$~FPS resulting in 34k frames, 
in addition to the previous 76k frames of ScanNet.
\tabref{tab:more_data} shows an impressive performance boost of $\mathbf{+2.0}$~AP$_{50}$.
For future work, considering that this improvement is obtained simply by adding more automatically generated masks to the pre-training, our approach seems promising to train on even more data.
It could even be plausible to train on internet images combined with monocular depth estimation, such as ZoeDepth~\cite{bhat2023zoedepth}, to compute the partial point clouds.

\begin{table}[!h]
    \centering
    \small
    \setlength{\tabcolsep}{0.1cm}
    \resizebox{0.48\textwidth}{!}{
        
\begin{tabular}{l|lll}
\toprule
     \textbf{Pre-Training Dataset} (\# Frames) & \textbf{AP} & \textbf{AP$_{50}$}\\
     \midrule
     ScanNet ($76$k)             & $12.0$ & $22.7$\\ %
     ScanNet ($76$k), ScanNet++ ($34$k) &
     $13.7$ \textcolor{darkgreen}{\footnotesize{(+\,$1.7$)}} &
     $24.7$ \textcolor{darkgreen}{\footnotesize{(+\,$2.0$)}} \\
     \bottomrule
\end{tabular}
}
    \caption{
        \textbf{Performance Increase with Additional Training Data.}
        }
        \vspace{-6px}
\label{tab:more_data}
\end{table}

\subsection{Qualitative Results}
\label{sec:quali}
Fig.\,\ref{fig:quali} shows several representative examples of \name{} segmentation results on the ScanNet++ dataset.
As can be seen, the scenes are quite diverse, presenting multiple challenges such as clutter and a wide range of mask sizes.
Despite these challenges, our model predicts quite accurate and well-localized segmentation masks. 
For example, compared to Mask3D,
our method is able to segment the finer-grained objects on top of the bed and the shelf.
The masks of the computer screen and the chair in front of it are also less fragmented than predictions of the baselines.
See the appendix (\secref{sec:wild}) for visualizations in the wild test.

\subsection{Application: Open-Set Scene Understanding}
\label{sec:openset}
A real-world application of our class-agnostic 3D segmentation method is open-vocabulary 3D scene understanding, as implemented by the recent OpenMask3D~\cite{Takmaz2023NIPS}.
Given a 3D scene, a user can search for arbitrary objects via text prompts (see Fig.~\ref{fig:openmask3d}).
A core component of OpenMask3D is Mask3D which segments the scene into masks.
Since Mask3D is trained on the closed-set of ScanNet, its masks are not truly class-agnostic or open-vocabulary.
We therefore replace Mask3D with our class-agnostic \name{}.
We evaluate on the closed-set labels of Replica and ScanNet++ in Table~\ref{tab:openmask3d}
and report an improvement of up to $\mathbf{+2.7}$~AP$_{50}$. Yet, this score serves only as an indicator, since most masks are not considered in this closed-set evaluation. 
See the appendix (\secref{sec:open_set}) for a discussion and more details.

\begin{table}[!h]
    \centering
    \small
    \setlength{\tabcolsep}{0.1cm}
    \resizebox{0.48\textwidth}{!}{
        \begin{tabular}{l|lll}
\toprule
     \textbf{Model} & \textbf{Segmentor} & \textbf{ScanNet++} & \textbf{Replica}\\
     \midrule
     OpenMask3D~\cite{Takmaz2023NIPS} & Mask3D\cite{Schult2023ICRA} & $15.0$ & $18.0$\\
     OpenMask3D~\cite{Takmaz2023NIPS} & \name{} (Ours) &
     $17.7$ \textcolor{darkgreen}{\footnotesize{(+\,$2.7$)}} &
     $18.7$ \textcolor{darkgreen}{\footnotesize{(+\,$0.7$)}} \\
     \bottomrule
\end{tabular}

}
    \caption{
        \textbf{Open-Set 3D Scene Understanding Scores.}
        The evaluation metric is average precision with an IoU threshold of $50$\%.}
        \vspace{-5px}
\label{tab:openmask3d}
\end{table}

\section{Conclusion and Discussion}
We have presented \name{}, a simple, yet powerful class-agnostic 3D segmentation model.
The model is trained entirely on automatically generated masks from SAM,
a foundation model for image segmentation.
It is competitive and even outperforms existing 3D segmentation models that rely on hand-labeled 3D training scenes.

Indeed, this raises the question of whether hand-labeled 3D training datasets are as essential as they were once thought to be.
Similarly, in the case of test labels, our qualitative comparison indicates that \name{} sporadically identifies small object masks not annotated in the ground truth.
Consequently, the full performance of \name{} might not be accurately reflected in the scores.

Overall, the work shows the potential of foundation models as automatic training label generators, and our preliminary experiments seem to indicate that more data will further increase the segmentation performance for \name{}.

{\small
\bibliographystyle{ieee_fullname}
\bibliography{main}
}
\clearpage
\setcounter{section}{0}
\setcounter{figure}{0}
\setcounter{table}{0}
\renewcommand\thesection{\Alph{section}}
\renewcommand\thetable{\Alph{table}}
\renewcommand\thefigure{\Alph{figure}}

\section{Implementation Details}
\label{sec:implement}
The feature backbone of \name{} is a Minkowski Res16UNet34C~\cite{Choy2019CVPR}.
We perform standard data augmentations, including horizontal flipping, random rotations, elastic distortion and random scaling.
In addition, we use color augmentations including jittering, brightness and contrast augmentation.
For Stage\,1, we use AdamW optimizer and a one-cycle learning rate schedule with a peak learning rate of $2 \times 10^{-4}$.
The model is trained for 20 epochs with a batch size of $16$ partial RGB-D point clouds.
A training on $2$\,cm voxelization takes approximately $60$ hours with $2$ RTX3090 GPUs.
For Stage\,2, the initial learning rate is set to $1 \times 10^{-4}$.
We train the model for 50 epochs with a batch size of $8$ full 3D point clouds.
Training takes $\sim 10$ hours with $2$\,cm voxels on $4$ A100 GPUs.
For both Stage\,1 and Stage\,2, we set the number of queries to 150 during training. For Stage\,1, following Mask3D~\cite{Schult2023ICRA}, the values of $\lambda_{\text {obj}}$, $\lambda_{\text {dice}}$, and $\lambda_{\text {ce}}$ are set to 2, 2, and 5, respectively. For Stage\,2, the values of $\lambda_{\text {dice}}$ and $\lambda_{\text {ce}}$ are set to 2 and 5, respectively.

In Stage\,1, to generate masks with SAM~\cite{Kirillov2023ICCV}, we use its automatic mask generation pipeline. We prompt SAM with a regular grid on the RGB image. After post-processing, we obtain nested masks with different levels of granularity. In cases where a pixel is encompassed by multiple masks, we assign the mask ID associated with the highest predicted Intersection over Union (IoU) to that pixel. Then we project the masks to 3D. In Stage\,2, we use DBSCAN~\cite{ester1996density} to split erroneously merged instances for the selected high-confidence masks.

\section{Additional Datasets}
\label{sec:wild}
\vspace{-5pt}
To further assess the generalization capability of \name{}, we test on additional indoor and outdoor data. 

\subsection{Indoor In-the-Wild Scenes}
\label{sec:indoor}
\begin{figure*}
    \centering
    \includegraphics[width=0.98\textwidth]{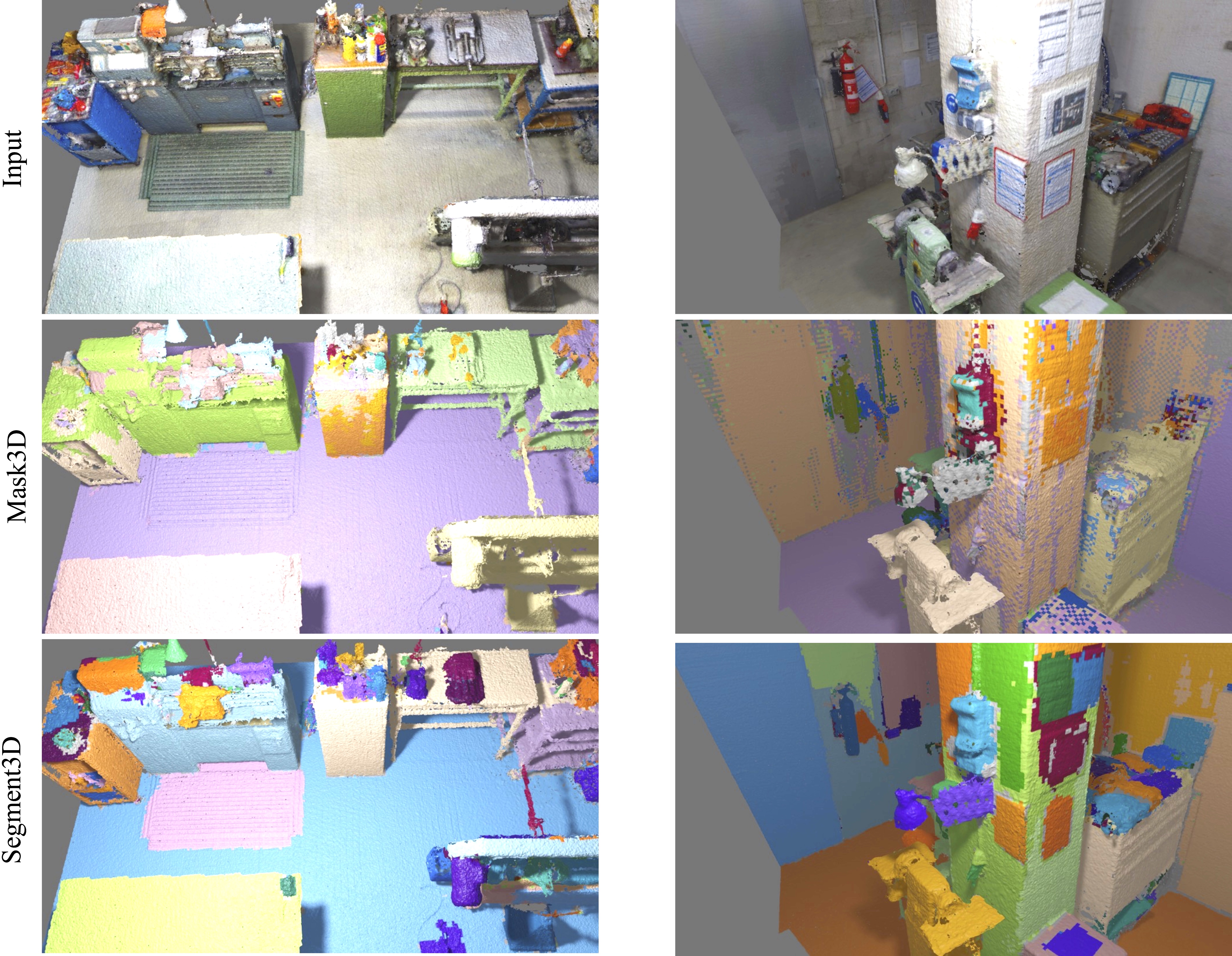}
    \caption{\textbf{Qualitative Results on Indoor \emph{`in-the-wild'} Scenes.}
    From top to bottom, we show the colored input 3D scenes,
    the segmentation masks predicted by Mask3D~\cite{Schult2023ICRA} and our Segment3D.
    Segment3D, trained with unsupervised learning, demonstrates superior generalization compared to Mask3D, which is trained with fully supervised learning.
    }
    \label{fig:wild}
\end{figure*}

\vspace{-5pt}
We first perform experiments on in-the-wild indoor scans.
Unlike the training scenes, which are mostly collected in hotel rooms, office spaces and universities, the scans used in these experiments are recorded from workshops depicting a wide variety of objects never seen during training, such as milling machines, saws, turning lathes, grindstones \etc{}, (see Fig.~\ref{fig:wild}).
We show qualitative results of our \name{} and Mask3D~\cite{Schult2023ICRA} which are both pre-trained on ScanNet~\cite{Dai2017CVPR}.
Our \name{} yields much better masks than Mask3D.
For example, \name{} yields accurate, sharp and compact segmentation masks of small individual objects
on the table, and posters on the wall while at the same time being able to segment large tools.

\subsection{Outdoor Scenarios}
\label{sec:outdoor}
\begin{figure*}
    \centering
    \includegraphics[width=\textwidth]{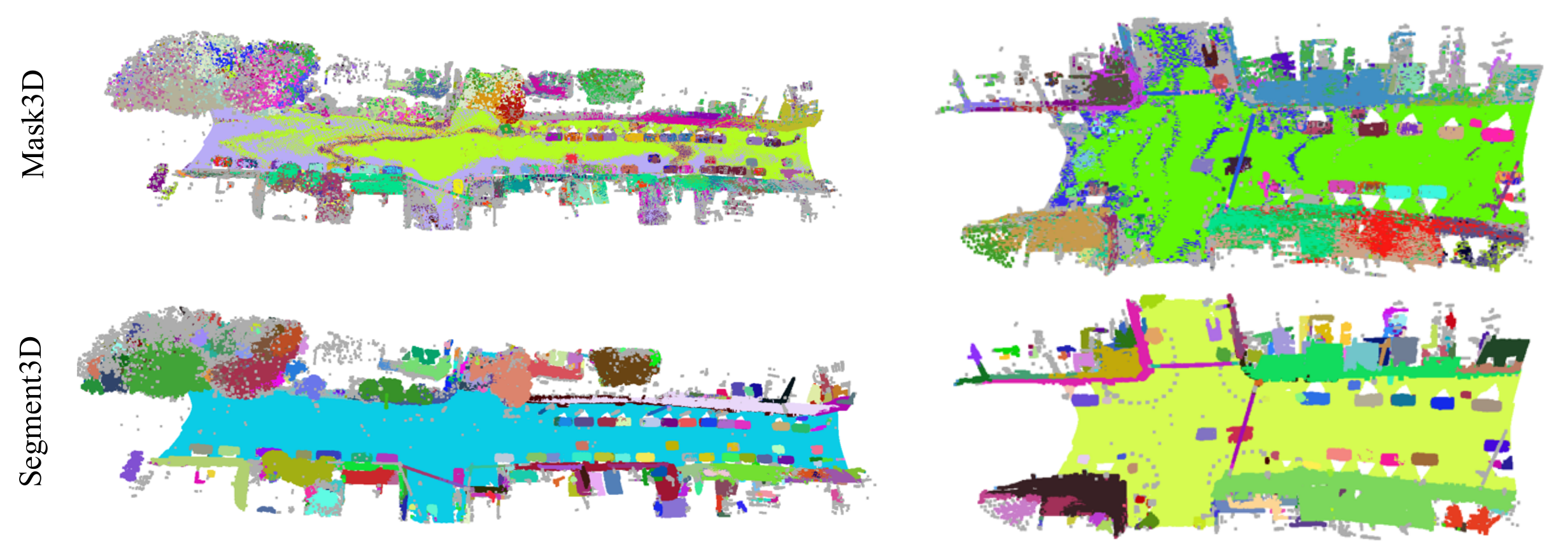}
    \caption{\textbf{Qualitative Results on Outdoor Scenes.}
    We show segmentations predicted by Mask3D~\cite{Schult2023ICRA} \emph{(top)} and our Segment3D \emph{(bottom).}
    }
    \label{fig:outdoor}
\end{figure*}

\vspace{-5pt}
Next, we visualize the performance of \name{} and Mask3D on outdoor scenes in Fig.~\ref{fig:outdoor} and report scores in Tab.~\ref{tab:outdoor}.
Specifically, we carry out experiments on the Paris-Lille-3D~\cite{roynard2018paris} (PL3D) dataset.
PL3D is a segmentation dataset containing more than $2$~km of street recordings of the French cities, Lille and Paris.
The test set labels are not publicly available.
However, since both Mask3D~\cite{Schult2023ICRA} and \name{} are trained on ScanNet, we can use the entire training set of the PL3D dataset for evaluating the effectiveness of class-agnostic segmentation.
We directly apply both models trained on room-sized indoor scenes to the outdoor scenes.
The only adaptation we perform is the split of the large outdoor scenes into multiple smaller crops, due to GPU memory limitations, and
the point cloud coordinates are scaled down by a factor of $10$ to be better aligned with the room-sized training data. %

The qualitative comparison reveals impressive generalization performance of \name{}, for example, the individual cars along the street are correctly segmented while Mask3D oversegments or merges them.
The street itself is correctly recognized as a very large single segment,
while Mask3D shows noisy artifacts.
Due to the necessary preprocessing and voxelization, the quantitative scores are low as many small ground truth segments are merged in a single voxel.
However, it can still be observed that \name{} performs significantly better than Mask3D.
This indicates that the generalization of \name{} is also better when transferring from indoor to outdoor scenes.

\begin{table}[!b]
    \centering
    \small
    \setlength{\tabcolsep}{0.1cm}
    \resizebox{0.49\textwidth}{!}{
        \begin{tabular}{l|lll}
\toprule
     \textbf{Model}& \textbf{AP$_{50}$} & \textbf{AP$_{25}$} \\
     \midrule
     Mask3D (Pre-trained on ScanNet)& $1.5$ & $3.3$ \\
     Segment3D (Pre-trained on ScanNet)  &
     $\mathbf{4.0}$ \textcolor{darkgreen}{\footnotesize{(+\,$2.5$)}} &
     $\mathbf{7.7}$ \textcolor{darkgreen}{\footnotesize{(+\,$4.4$)}} \\
     \bottomrule
\end{tabular}}
    \vspace{3px}
\caption{
        \textbf{Evaluation on the Paris-Lille-3D Dataset.}}
\label{tab:outdoor}
\end{table}

\section{Open-Set 3D Scene Understanding}
\label{sec:open_set}
\begin{figure*}
    \centering
\definecolor{amethyst}{rgb}{0.6, 0.4, 0.8}
\begin{overpic}[width=0.9\textwidth, ]{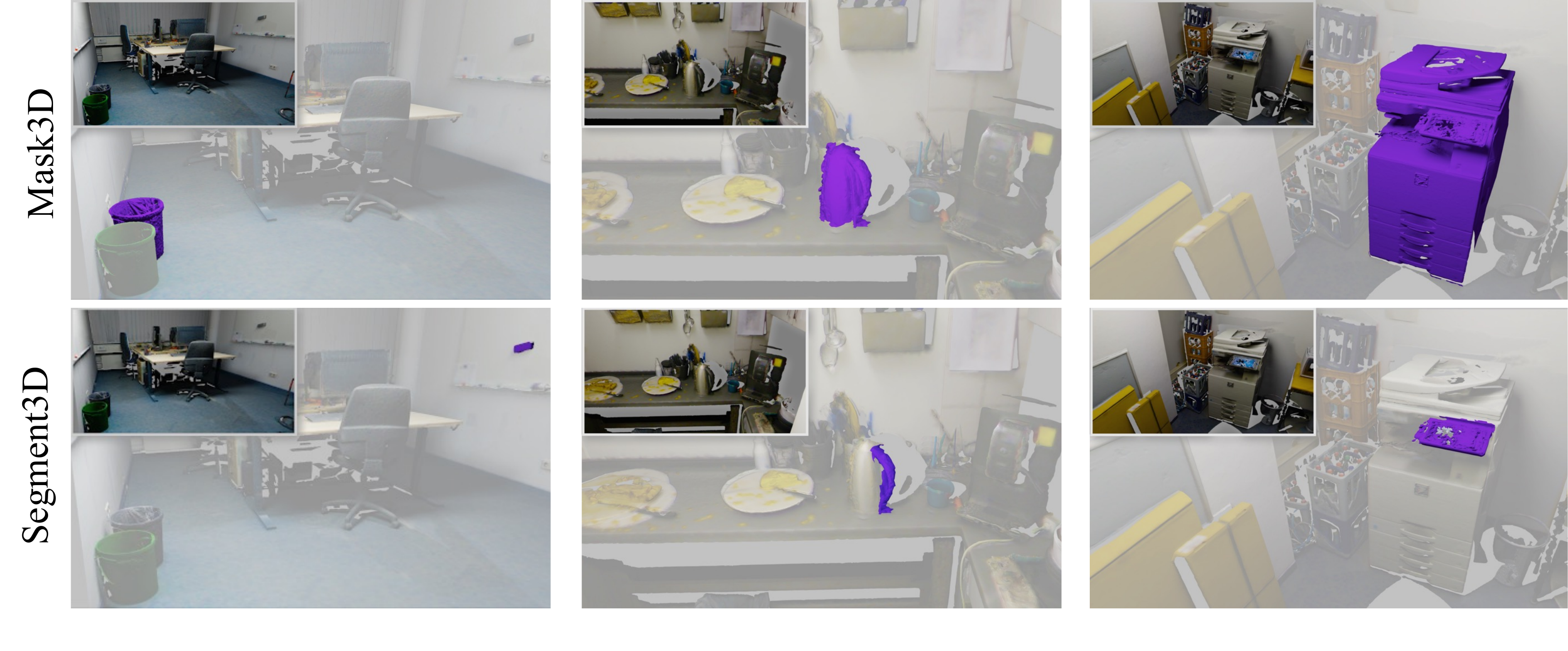}%
    \put(12,0.5){{\small \colorbox{white}
    {\textit{``a black eraser''}}}}
    \put(45,0.5){{\small   \colorbox{white}{\textit{``kettle handle"}}}}
    \put(74,0.5){{\small   \colorbox{white}{\textit{``copier control screen"}}}}
\end{overpic}
    \caption{\textbf{Qualitative Results of Adapted OpenMask3D~\cite{Takmaz2023NIPS}.}
    Given a text prompt \emph{(bottom)}, OpenMask3D finds the corresponding masks~\newcolorsquare{amethyst}{amethyst} in a given 3D scene \emph{(top)}. We show the 3D scene reconstruction and an RGB image for better visualization \emph{(top left corner)}.
    The original OpenMask3D based on Mask3D is unable to recognize any of the above queries, whether it is a relatively small object or a fine-grained affordance part of an object. In contrast, our Segment3D adaptation performs well in these cases.
    }
    \label{fig:supp_openmask3d}
\end{figure*}

\vspace{-5pt}
As mentioned in the main paper, a key application of our class-agnostic segmentation is open-set 3D scene understanding where a user can search for arbitrary open-vocabulary descriptions. 
This task is implemented in the OpenMask3D \cite{Takmaz2023NIPS} framework. 
While the original OpenMask3D relies on Mask3D \cite{Schult2023ICRA} as the scene segmentor, we adapt it and replace Mask3D with \name{}. Below, in Fig.~\ref{fig:supp_openmask3d} we show additional results of the original OpenMask3D based on Mask3D, compared to our adapted OpenMask3D based on Segment3D.

In addition to performing fine-grained segmentation, \name{} also offers more flexible input for text prompts, such as the affordance part of an object.
In Fig.~\ref{fig:supp_openmask3d}, we provide two examples.
When we refer to the ``kettle handle", Segment3D can highlight it accurately while Mask3D only segments the whole ``kettle".
Similarly, Segment3D can point out the ``copier control screen" while Mask3D can only segment the whole ``copier". This enables new applications for intelligent non-human assistants, autonomous robots, and AR/VR devices.

\end{document}